\documentclass{article}

\pdfoutput=1

\usepackage[final, nonatbib]{neurips_2022}

\usepackage{amsmath,amssymb} 
\usepackage{graphicx}
\usepackage{float}

\usepackage[utf8]{inputenc} 
\usepackage[T1]{fontenc}    
\usepackage{url}            
\usepackage{booktabs}       
\usepackage{amsfonts}       
\usepackage{nicefrac}       
\usepackage{microtype}      
\usepackage{xcolor}         

\title{Intentional Choreography with Semi-Supervised Recurrent VAEs}

\author{%
  Mathilde Papillon \thanks{Corresponding email: papillon@ucsb.edu} \\
  UC
  Santa Barbara\\
  \And
  Mariel Pettee \\
  Lawrence Berkeley
  National Lab \\
  \And
  Nina Miolane \\
  UC
  Santa Barbara \\
}

\begin{document}

\maketitle


\begin{figure}[H]
	\centering
 	\includegraphics[width=1.\linewidth]{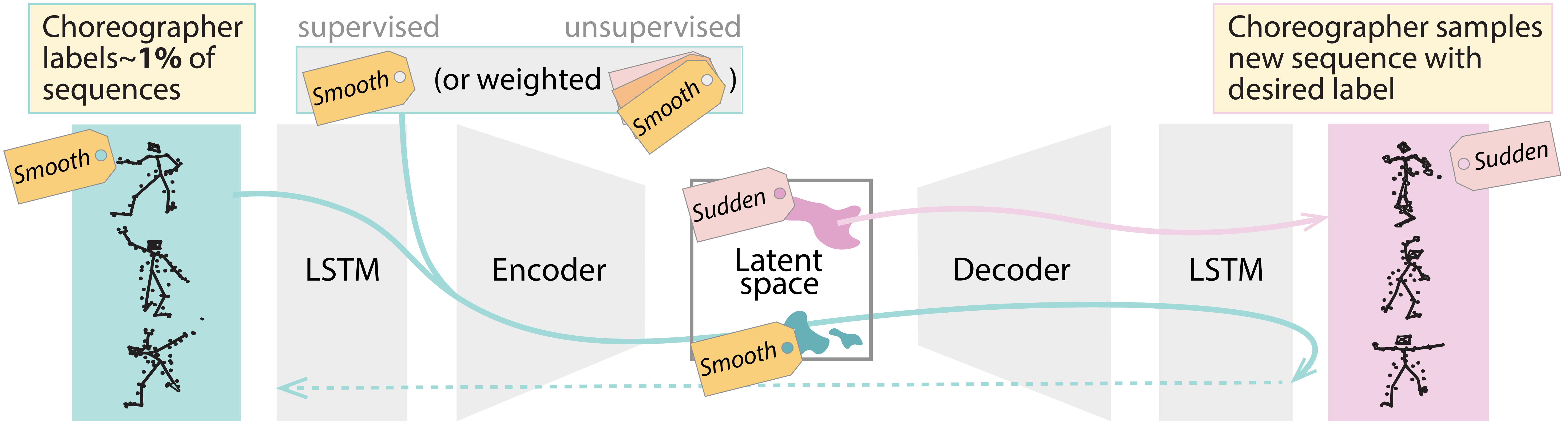}
	\caption{\textsc{PirouNet}, a semi-supervised recurrent variational autoencoder, is trained (blue path) with a 1\% manually labeled dataset. New dance sequences are conditionally generated (pink path) by sampling in the latent space conditionally on the desired choreographic label.}
	\label{fig:model}
\end{figure}

\section{Introduction}

Artificial intelligence (AI) has given rise to a suite of methods that automatically generate new dance sequences. Yet, these tools continue to be largely ignored by dance practitioners because they lack the ability to create dance sequences with specific choreographic aesthetic. Two broad categories describe how these AI methods engage with an artist's choreographic practice. The first consists of deep learning approaches that often rely on recurrent networks \cite{og_rnns} and (variational) autoencoders \cite{Kingma2014vae}. These act as generators for new movement ideas, either by randomly generating sequences \cite{berman_from_scratch_2015,from_scratch_li,augello_2017,pettee_2019} or by responding to a specific movement \cite{augello_2017,pettee_2019,james_2018_autoencoder_live} or music prompt \cite{aist_music,lee_2019,alemi_2017,from_music_transformer}. Yet, the user’s creative control over these sequences is restricted to the choice of training data or of an external prompt. The second category, made up of algorithmic and non-deep methods \cite{laban_editor_AI, zhang_li_2006}, makes use of Laban Movement Analysis (LMA) \cite{laban_book1,groff1995laban}, a widely recognized dance theory \cite{laban_mvt_dance_concepts}, to translate choreographic scores directly into movement. Beyond providing a static starting point for inspiration \cite{catalyst}, this category's methods are limited in their creative contribution. We postulate that the lack of dance generation methods with significant creative control comes from a lack of large, user-specific dance datasets with labels meaningful to the user's practice. Such datasets for daily human actions \cite{ntu_dataset} have enabled supervised action generation methods \cite{action_2_motion,actor_vae}. By contrast, the annotated dance databases that do exist \cite{aist_database,kpop_database,web_app_annotations} are small and limited to their producers' specific styles and creative processes.

To remedy this, we propose \textsc{PirouNet}, a semi-supervised generative recurrent deep learning model that conditionally creates new dance sequences from choreographers’ aesthetic inputs. Our semi-supervised approach, combined with a suite of tools for automatic labeling, ensures that the choreographer only needs to label a very small portion of an input dataset. While we use the categorical intensities of LMA's Laban Time Effort \cite{laban_book2_2006,def_efforts,robot_efforts} as an illustrative example, \textsc{PirouNet} users can implement any choice of subjective labels they wish to use.

\section{Methods}

\textsc{PirouNet} (github.com/bioshape-lab/pirounet) features a dance encoding and generative model that uses (i) a variational autoencoder (VAE) \cite{Kingma2014vae} inspired from \cite{pettee_2019}, coupled with (ii) motion dynamics through a long-short-term-memory network \cite{Hochreiter97}. For semi-supervised training \cite{Kingma2014dgm}, \textsc{PirouNet} leverages iii) a linear classifier. We propose the following conditional dance generative model: $p_\pi(y)
= \operatorname{Cat}(y \mid \pi); \quad p(z)=\mathcal{N}(z \mid 0, I) ; \quad p_{\theta}(x \mid y, z; \theta)=\mathcal{N}(f_\theta(y, z), \sigma^2)$.
Here, $z$ is a continuous variable representing the dynamics, or \textit{which} movement is performed. $y$ is a categorical variable denoting the choreographer's label, or \emph{how} the move is performed. PirouNet solves the inverse problem of inferring the marginally independent latent variables ($z, y$) from a sequence ($x$). 

Implementation-wise, Fig \ref{fig:model}'s blue path denotes the flow of information during training. In the supervised case, \textsc{PirouNet} maximizes the log-likelihood $\log p_\theta(x, y)$ via the maximization of its lower bound $-\mathcal{L}(x, y)$, which resembles a VAE's regularized $L_2$ reconstruction loss (see App. 3). In the unsupervised case, $y$ is missing and treated as another latent variable, in addition to $z$, over which we perform posterior inference. We maximize the marginalized log-likelihood $\log p_\theta (x)$ via the maximization of its lower bound (see App. 4). This consists of an entropy term and $\mathcal{L}(x, y)$ weighted by the classifier's confidence associated to each label $y$. Fig. \ref{fig:model}'s pink path shows how \textsc{PirouNet} conditionally generates dance. A new random latent variable $z$, representing body motion, is sampled from an approximation of the marginal distribution $q_\phi(z|y)$ where the label $y$ is chosen by the user.

\section{Results}
\begin{figure}[ht]
	\centering
 	\includegraphics[width=0.9\linewidth]{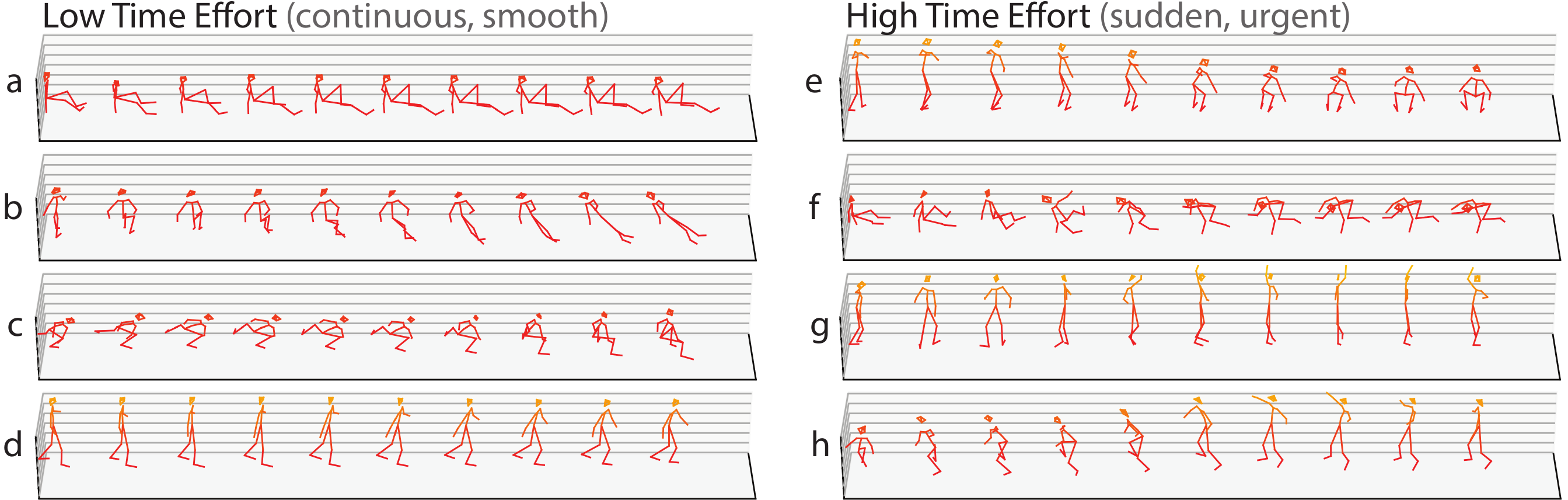}
	\caption{New sequences generated by \textsc{PirouNet}, demonstrating its dynamic creativity in adequation with the choreographer's desired labels. a. Slow leg extension from V-sit. b. Continuous transition from crouched to plank position. c. Smooth weight transfer through deep plié. d. Forward weight transfer through small plié. e. Spontaneous drop to a crouched position. f. Leg swing leads sharp torso rotation. g. Pirouette with arm thrown up. h. Explosive extension of torso from plié.}
	\label{fig:results}
\end{figure}

After an extensive hyperparameter search (see App. 1 and 2), we validate \textsc{PirouNet} on a 225-sequence benchmark set (75 per label), some examples of which are shown in Fig. \ref{fig:results}. The labeler identifies 96.0\% of these sequences to be realistic and novel. Of these, 63\% are determined to be in agreement with their intended label, or 83\% of the labeler's self-agreement upon labeling the dataset twice. Further, \textsc{PirouNet} originals feature a diversity metric \cite{action_2_motion} more than double that of test data. These results show \textsc{PirouNet}'s ability to perform as a creative tool, even with limited supervision.

\section{Context within the creative practice}
We hope this artist-guided, adaptable tool will act as a launching pad for engaging with previous work and developing vocabulary for new choreography, offering new insight into one's own choreographic vision. In parallel, identifying qualitative intentions on a spectrum (low to high, small to big, etc) offers paths towards exploring movement that contrasts with one's usual practice. Moreover, \textsc{PirouNet} helps facilitate a conversation between old repertoire and new explorations. The artist is empowered to shape their AI tool by a customizable keypoint labeling web-app that supports graphic-supported classification of any keypoint dataset with any annotations (see App. 5). While demonstrated on dance, the proposed method can be extended to other forms of art creation, inspiring AI-based tools tailored to the style and intuition of their artist.

\pagebreak
\section{Ethics}
All training data was sourced ethically, originating from our second author. As the dataset only represents one body and one movement practice, it cannot validate \textsc{PirouNet} for all dancers or practices. Moreover, it is important to underline that LMA was designed for and mostly used by the Western modern dance community, meaning the results presented should not be generalized to all forms of dance annotations. The open sourced nature of the project and low data barrier were designed to encourage many more dance practitioners to train their own version of \textsc{PirouNet}, with or without LMA labels.

\bibliographystyle{unsrt}
\bibliography{main}

\appendix

\section{Appendix}

\subsection{Datasets}
Like most deep learning tools for dance, we leverage motion data in keypoint format. This format, often included in large movement datasets \cite{aist_database,ntu_dataset,hdm05_database}, represents the body as a cloud of 3D points, each representing a unique joint. Available and high-performing pose-estimation software \cite{deeppose,stacked_pose,multi_pose} makes keypoint format accessible to smaller, homemade datasets as well. In this spirit, we use half of Pettee's keypoint dataset \cite{pettee_2019}.

The dataset \cite{pettee_2019} is comprised of 36,396 poses extracted from six uninterrupted dances captured at a rate of 35 frames per second. This amounts to about 20 minutes of real-time movement of an experienced dancer. Each pose features 53 joints captured in three dimensions, normalized such that the dance fits within inside a unit box. The dancer's barycenter is fixed to one point on the (x,y) plane. From the pose data, we extract 36,356 sliding sequences of 40 continuous poses, and manually label 350 of these sequences (0.96\% of the dataset) which do not share any of the same poses. This takes an experienced dancer (the principal author) about 3 hours, identifying if the movement's Laban Time Effort (characterizing how sudden or urgent a movement feels) is Low, Medium, or High. Using our label augmentation toolkit (see App. 3), we apply two techniques to get 9,167 labeled sequences (representing 25.2\% of our unlabeled dataset) in total, split between 45\% Low, 34\% Medium, and 21\% High Efforts. (i) We automatically label all sequences between sequences that share a same Effort. For example, if two back-to-back sequences are deemed to have Low Time Effort, all sequences that are a combination of the poses in these two sequences are also labeled with Low Time effort. (ii) We extend every label to all sequences starting within 6 frames (0.17 seconds) before or after its respective sequence.

\subsection{Training}
PirouNet is built using the PyTorch library \cite{pytorch} and run on a server with two Nvidia A30 GPUs and two CPUs, each with 16 cores. We train using an Adam optimizer with standard hyperparameters \cite{adam_opt}. We present results for the PirouNet architecture resulting from a hyperparameter search using Wandb \cite{wandb} on batch size, learning rate, number of LSTM and dense layers, as well as hidden variable sizes. PirouNet uses 5 LSTM layers with 100 nodes in both the encoder and the decoder. The classifier features 2 ReLU-activated \cite{relu} linear layers with 100 nodes. The latent space is 256-dimensional, which is approximately 25 times smaller than the 6360-dimensional initial space. We train for 500 epochs with a learning rate of $3\text{e}^{-4}$ and a batch size of 80 sequences. We select different epochs for PirouDance and PirouWatch to minimize validation loss and optimize certain evaluation metrics presented in the appendices. For unsupervised training, we use 35,538 40-pose sequences, with the remaining sequences being reserved for testing. For supervised training, PirouNet\textsubscript{dance} and PirouNet\textsubscript{watch} are trained on 79\% (16.6\% of entire training set) and 92\% (18.8\% of entire training set) of the labeled sequences, respectively. We reserve 5\% of the labeled sequences for validation, and 3\% for testing.

\subsection{Derivation of Lower Bound in the Labeled Case}
This appendix presents a step-by-step derivation for the evidence lower bound (ELBO) used for training PirouNet on labeled input data. As shown below, the ELBO provides a computable quantity that is smaller or equal to the log-likelihood (which is, itself, intractable due to the integral on the hidden latent variables). The goal is to find the ``most likely" parameters of our model, i.e. the parameters that maximize the model's log-likelihood. Instead of maximizing the (intractable) log-likelihood, we will maximize its ELBO.

In the case of input data $x$ that have label $y$, the log-likelihood is $p(x, y)$ by definition. Let $q(z|x,y)$ be the approximate posterior for the continuous latent variable z. The law of total probability helps us write the log-likelihood integrating on the hidden latent variable $z$:
\begin{align*}
\log p(x, y) &= \log \int_z p(x, y, z) d z \\
&=\log \int_z q(z \mid x, y) \frac{p(x, y, z)}{q(z \mid x, y)} d z
\end{align*}

For a concave log, Jensen's inequality states that

\begin{align*}
\log \mathbb{E}_{q(z \mid x, y)} f(z) \geq \mathbb{E}_{q(z \mid x, y)} \log f(z),
\end{align*}

for any function $f$. Using this:
$$
\\
\Rightarrow \log p(x, y) \geq \int q(z \mid x, y) \log \left(\frac{p(x, y, z)}{q(z \mid x, y)}\right) d z
$$
The definition of conditional probability gives:  $p(x, y, z)=p(x \mid y, z) \cdot p(y, z) $. Assuming that $z$ (which movement is performed) and $y$ (how the movement is performed) are independent random variables, we get $p(y, z) = p(y)\cdot p(z)$ and thus: $p(x \mid y, z) \cdot p(y) \cdot p(z)$,

\begin{align*}
&\Rightarrow \log p(x, y) \geq \int_z q(z \mid x, y) \log \left(\frac{p(x \mid y, z) \cdot p(y) \cdot p(z)}{q(z \mid x, y)}\right) d z \\
&=\int_z q(z \mid x, y)[\log p(x \mid y, z)+\log p(y)] d z+\int_z q(z \mid x, y) \log \left(\frac{p(z)}{q(z \mid x, y)}\right) d z \\
&\text{ By definition of the KL-divergence: $KL(q\parallel p) = \int q \log \left(\frac{q}{p}\right)$ \cite{kldiv}:} \\
&=\int_z q(z \mid x, y)[\log p(x \mid y, z)+\log p(y)] d z-K L(q(z \mid x, y) \| p(z)) \\
&=\mathbb{E}_{q(z \mid x, y)}[\log p(x \mid y, z)+\log p(y)]-K L(q(z \mid x, y) \| p(z)) \\
&=-\mathcal{L}(x, y)
\end{align*}

What does the first term represent?
Invoking the decoder generative model ${\text{Dec}}_{\theta}$ for some input $x_{i}$, we get :
\begin{align*}
x_{i}&=\operatorname{Dec}_{\theta}\left(z_{i}, y_{i}\right)+\varepsilon_{i} \text {, for some } \varepsilon_{i} \sim \mathcal{N}(0,\sigma^2)\\
\text{Therefore:} \\
\log p(x \mid y, z) &=\log \mathcal{N}\left(\operatorname{Dec}_{\theta}(z, y), \sigma^2\right) \\
&=\log \frac{1}{\sqrt{2 \pi \sigma^{2d}}} \exp \left[-\frac{\left\|x_{i}-\operatorname{Dec}_{\theta}\left(z_{i}, y_{i}\right)\right\|^{2}}{2\sigma^2}\right] \\
&=\log \left(\frac{1}{\sqrt{2 \pi\sigma^{2d}}}\right)-\frac{\left\|x_{i}-\hat{x}_{i}\right\|^{2}}{2\sigma^2}
\end{align*}
Dropping the constants (shifting and scaling), we see that this is just the plain absolute difference squared between the input $x_i$ and the VAE's reconstructed output $\hat{x}_{i}$. Maximizing the ELBO $\mathcal{L}(x, y)$ means maximizing, in part, $\log p(x \mid y, z)$ and thus minimizing the reconstruction loss $\left\|x_{i}-\hat{x}_{i}\right\|^{2}$. 

The second and third terms represent a regularization with respect to this reconstruction loss. 

\subsection{Derivation of Lower Bound in the Unlabeled Case}
In this section we walk through the derivation for obtaining the evidence lower bound (ELBO) used for training on unlabeled data. In this case, we seek to find an expression smaller or equal to the log-likelihood $\log p(x)$ over unlabeled inputs $x$. We will make use of a uniform prior over all categorical labels $y$, as well as a normalized probability distribution over the continuous latent variable $z$.

Applying the law of total probability to the discrete probability distribution for categorical labels $y$, and then to the continuous probability distribution over latent variable $z$, we get:

\begin{align*}
\log p(x) &=\log \sum_{y} p(x, y) \\
&=\log \sum_{y} \int_z p(x, y, z) d z \\
&=\log \sum_{y} \int_z q(z, y \mid x) \frac{p(x, y, z)}{q(z, y \mid x)}d z.
\end{align*}

By the definition of conditional probability $q(z, y \mid x) = q(y \mid x) \cdot q(z \mid x, y)$:
\begin{align*}
\log p(x)=\log \sum_{y} q(y \mid x) \int_z q(z \mid x, y) \frac{p(x, y, z)}{q(z, y \mid x)} d z
\end{align*}
Invoking Jensen's inequality stating that $\log \mathbb{E}_{q(y \mid x)} f(y) \geqslant \mathbb{E}_{q(y \mid x)} \log f(y)$ for any function $f$:

\begin{align*}
\log p(x) \geqslant \sum_{y} q(y \mid x) \log \int_z q(z \mid x, y) \frac{p(x, y, z)}{q(z, y \mid x)} d z.
\end{align*}

Invoking Jensen's inequality again, as $\log \mathbb{E}_{q(z \mid x, y)} h(z) \geqslant \mathbb{E}_{q(z \mid x, y)} \log h(z)$ gives:
\begin{align*}
\Rightarrow \log p(x) &\geq \sum_{y} q(y \mid x) \int_z q(z \mid x, y) \log \left(\frac{p(x, y, z)}{q(z, y \mid x)}\right) d z. 
\end{align*}
Using the definition of conditional probability $q(z, y \mid x) = q(z \mid x, y) \cdot q(y\mid x)$, we get:
\begin{align}
&\Rightarrow \log p(x) \geq \sum_{y} q(y \mid x) \int_z q(z \mid x, y) \log \left(\frac{p(x, y, z)}{(q(z \mid x, y) \cdot q(y\mid y)}\right) d z\\
&=\sum_{y} q(y \mid x)\left(\int q(z \mid x, y) \log \left(\frac{p(x, y, z)}{q(z \mid x, y)}\right) d z-\log q(y\mid x)\right)\\
&=\sum_{y} q(y \mid x)(-\mathcal{L}(x, y)_{\text {unlabeled }}-\log q(y \mid x))\\
&=-\mathbb{E}_{q(y \mid x)} \mathcal{L}(x, y)_{\text {unlabeled }}+\mathcal{H}(q(y \mid x)),
\end{align}
where the last line uses the definition of the entropy $\mathcal{H}(q) = - \sum_y q(y) \log q(y)$, and $-\mathcal{L}(x, y)_{\text {unlabeled }}$ is the ELBO for the log-likelihood of the labeled case computed previously.

Our lower bound in this case includes the same regularized reconstruction loss $\mathcal{L}(x, y)_{\text {unlabeled }}$ as the labeled case, except now it is weighted by the encoding probability for each possible label $y$. This case also features a new term $\mathcal{H}$ which is defined as the entropy of the classification.

\subsection{Web-app for annotations}
\begin{figure}[ht]
	\centering
 	\includegraphics[width=\linewidth]{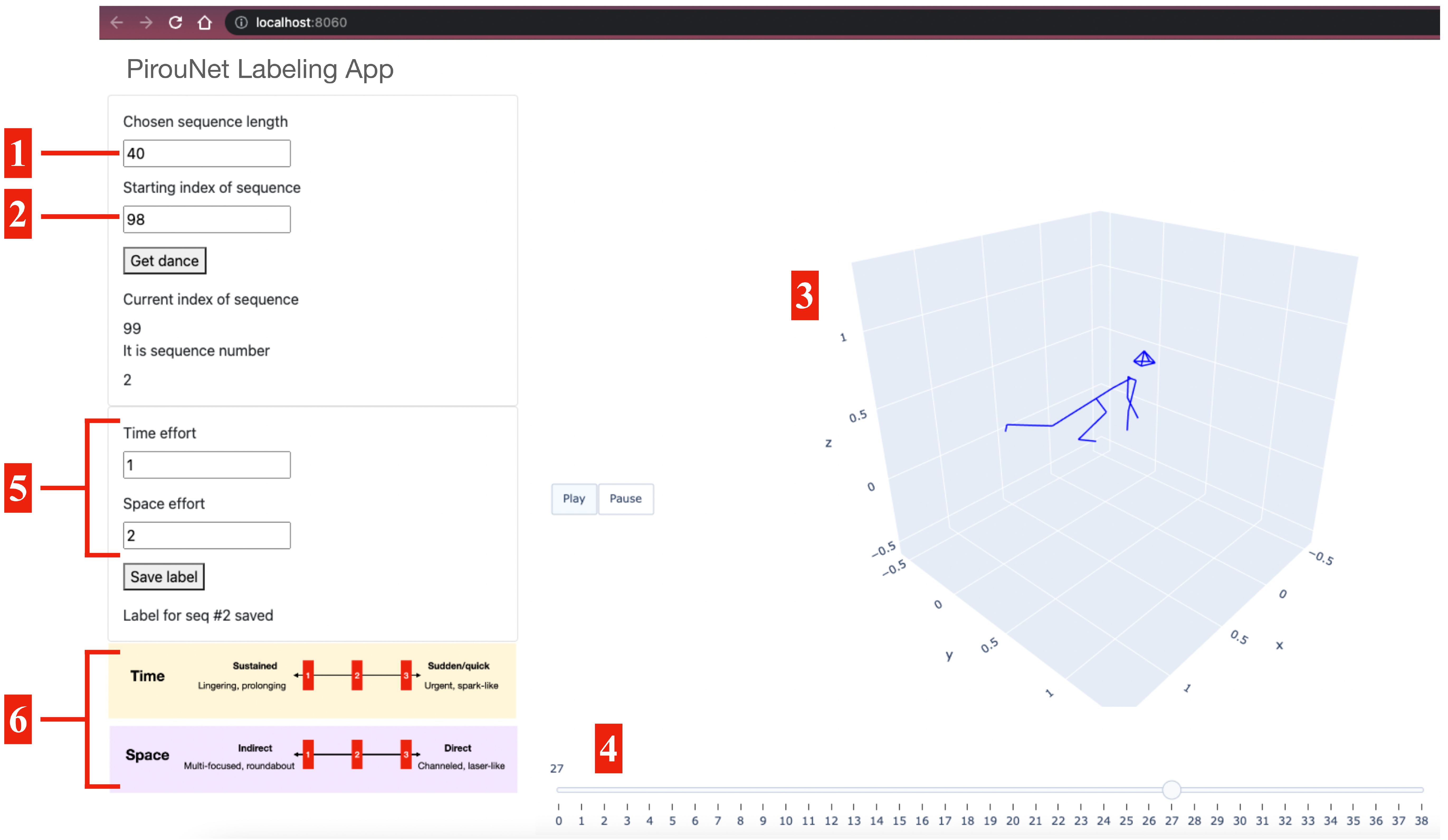}
	\caption{Screen capture of web labeling app. User enters amount of poses per sequence in (1), and the index of the starting pose of the first sequence in (2). Upon clicking ``Get Dance," an animation of the fully connected skeleton appears in (3). The user can zoom and rotate the animation directly, and click into specific frames with (4). User enters Laban Effort (or any chosen label) in (5) and clicks ``Save label" to record the inputs to a CSV. A spot is reserved in (6) for an infographic with instructions to encourage consistent labeling.}
	\label{fig:webapp}
\end{figure}

We propose a tool for easy manual labeling of an input dance dataset. This locally hosted Dash \cite{dash_plotly} app can be used by a choreographer wishing to classify their own movement. The graphical-user-interface, displayed in Fig. \ref{fig:webapp} is easy to navigate and allows multi-session labeling with a choice of starting pose index. The open source project is built such that plotting functions are easy to switch out and customize for any given keypoint dataset.

\end{document}